%% file: arxiv_sub.tex
\documentclass[10pt,twocolumn,a4paper]{article}
\usepackage[a4paper,left=19mm,right=19mm,top=25mm,bottom=25mm,columnsep=8mm]{geometry}
\usepackage[T1]{fontenc}
\usepackage{newtxtext,newtxmath}
\usepackage{graphicx,amsmath,bm,booktabs,multirow,caption}
\usepackage{cite}
\usepackage{url}
\usepackage[hidelinks]{hyperref}
\usepackage[capitalise,noabbrev]{cleveref}
\usepackage{titlesec}
\titleformat{\section}{\normalsize\bfseries}{\thesection.}{0.5em}{}
\titleformat{\subsection}{\normalsize\bfseries}{\thesubsection.}{0.5em}{}
\titlespacing*{\section}{0pt}{1.5ex}{1ex}

\providecommand{\vx}{}\renewcommand{\vx}{\bm{x}}
\providecommand{\vv}{}\renewcommand{\vv}{\bm{v}}
\providecommand{\vc}{}\renewcommand{\vc}{\bm{c}}

\makeatletter
\providecommand\abx@aux@refcontext[1]{}
\providecommand\abx@aux@cite[2]{}
\providecommand\abx@aux@segm[3]{}
\providecommand\abx@aux@defaultrefcontext[5]{}
\providecommand\abx@aux@defaultlabelprefix[2]{}
\makeatother

\title{\bfseries A Benchmark for Spatially Grounded Gesture Generation}
\author{Anna Deichler$^{1}$ \and Rishabh Dabral$^{2}$ \and
        Fethiye Irmak Dogan$^{3}$ \and Anindita Ghosh$^{2}$ \and
        Jonas Beskow$^{1}$ \\[2pt]
        {\small $^{1}$KTH Royal Institute of Technology \quad
                $^{2}$MPI for Informatics \quad
                $^{3}$University of Cambridge}}
\date{}

\begin{document}
\twocolumn[{%
  \maketitle
  \vspace{-1.5em}
  \begin{center}
  \begin{minipage}{0.86\textwidth}
    \small\noindent\textbf{Abstract.}\;
Communication in shared space interweaves verbal and non-verbal signals, and among them pointing gestures anchor language to the environment: ``put the cup on \emph{that} one'' is uninterpretable without the gesture that fixes the referent. For embodied agents to produce such gestures, their evaluation must ask whether a gesture indicates the intended object, yet no common framework for this exists; distributional metrics reward a gesture aimed at the wrong object as long as it looks natural. We introduce a
benchmark for spatially grounded gesture generation. It supplies three missing
pieces: a paired data release of ${\sim}$2K pointing-annotated clips from
naturalistic VR dialogue, each labelled with a ground-truth referent in a 3D
scene graph, alongside 1K scripted single-target clips with shared actors; a
task formulation in which systems receive spontaneous conversational speech and
a 3D target and must decide when to gesture, how to move, and how to aim, with
no oracle apex frame; and an evaluation protocol separating temporal alignment
from spatial grounding and reporting their product, complemented by a
crowdsourced naturalness study. We evaluate a flow-matching baseline, an
independently developed retrieval-based system, and a ground-truth reference
run. The retrieval-based system achieves substantially stronger spatial
grounding than both the baseline and captured motion, while being rated no more
natural than the baseline. Temporal evaluation likewise separates the systems,
but comparison with captured motion depends on the temporal scoring parameters.
These results expose a gap between geometric task metrics and perceived motion
quality, motivating evaluation that reports spatial, temporal, and perceptual
factors separately.
  \end{minipage}
  \end{center}
  \vspace{0.5em}
  \begin{center}
    \includegraphics[width=0.86\textwidth]{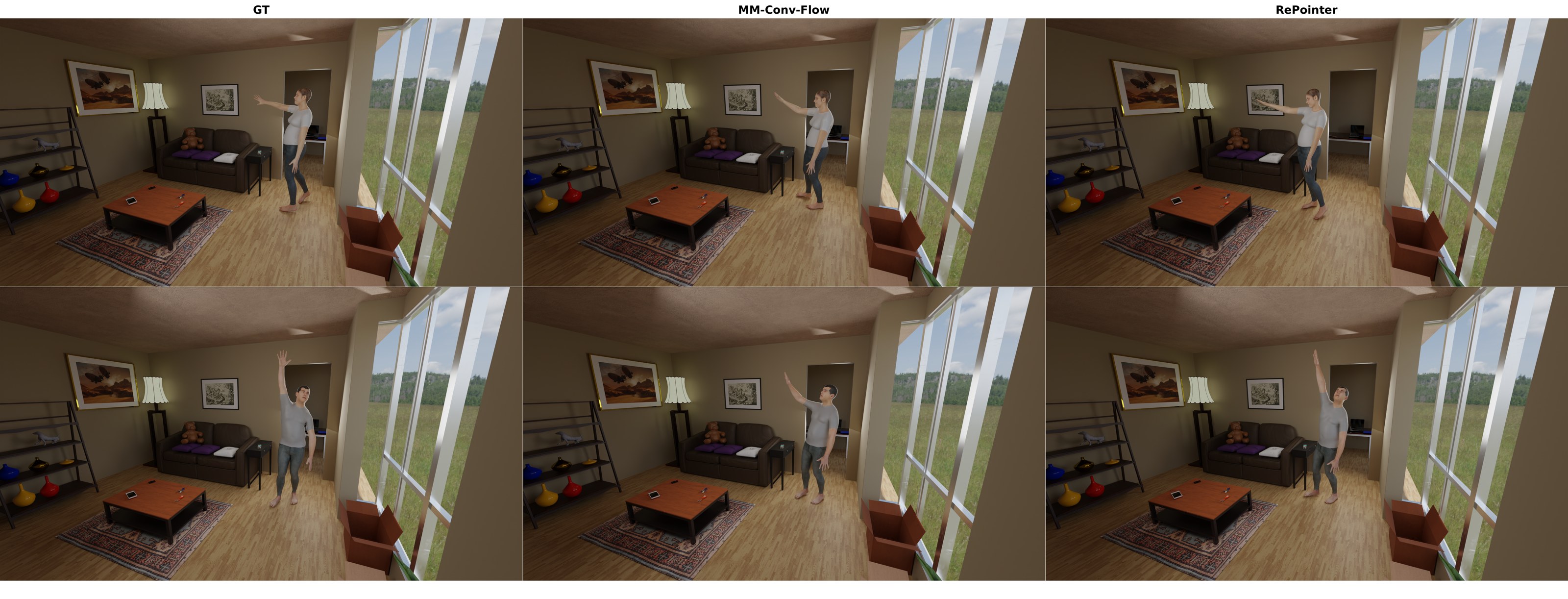}
    \captionof{figure}{Qualitative comparison of reference motion (GT),
      MM-Conv-Flow, and RePointer on two examples from the spatially grounded
      gesture benchmark.}
    \label{fig:teaser}
  \end{center}
  \vspace{1em}
}]

\input{body-arxiv.inc}

{\small
\bibliographystyle{unsrt}
\bibliography{egbibsample}}
   \clearpage

\input{appendix-arxiv.inc}

\end{document}

%% file: body-arxiv.inc
\section{Introduction}

Human communication in shared physical space runs on more than words. Speakers
maintain a representation of their surroundings and index it as they talk---through
stance and body orientation, through gaze, and through pointing---so that meaning
is distributed across the verbal and bodily channels rather than carried by either
alone. Reference to the environment is the clearest case: ``put the cup on
\emph{that} one'' is uninterpretable without the gesture that fixes what ``that
one'' denotes, and the gesture is uninterpretable without the room.

Embodied agents that share space with people---virtual avatars, social robots, AR
assistants---need both halves of this ability: to interpret such signals, and to
produce them. Our concern here is production. A gesture generation model can
produce fluent, speech-synchronised motion and still lack the capability that
makes communication in space work: indicating something about the scene.

Current co-speech gesture generation mostly addresses the distribution-matching
problem in motion synthesis. Models are trained to reproduce the statistical
properties of human motion and evaluated accordingly, with metrics such as
Fr\'echet Gesture Distance~\cite{yoon2020fgd} and diversity, as established by
the GENEA Challenge~\cite{kucherenko2023genea} for community-wide benchmarking.
For beat and iconic gestures, where the goal is natural-looking motion
synchronised with speech, this is appropriate. Referential gestures pose a
fundamentally different problem: a pointing gesture directed at the wrong object
can score well under distributional metrics while being a complete communicative
failure. Referential gestures need to be both natural and
referentially accurate. What has been missing is the data-driven counterpart: a
corpus, task, and metric set that let speech-conditioned generative models be
compared on referential accuracy the way they are already compared on motion
realism. Evaluating referential gestures therefore requires complementing
distributional fidelity with task success: did the gesture indicate the intended
object?

The shift has been blocked by three missing pieces: datasets pairing human
gesture with 3D scene context, a task formulation for spatially grounded
referential gesture generation, and metrics that measure spatial grounding in
gestures. We
address all three, building on MM-Conv~\cite{deichler2026mmconv}, whose
referring expressions are each annotated with a ground-truth referent in a 3D
scene graph, thereby enabling evaluation of whether generated motion geometrically indicates the intended object.

This paper defines the benchmark and validates its protocol using a
flow-matching baseline, an independently developed system (RePointer~\cite{he2026repointer}), and a
ground-truth reference run. Our contributions are:

\begin{enumerate}
\item A \textbf{paired data release}: ${\sim}$2K pointing-annotated clips from
  MM-Conv's naturalistic dialogue and 1K clean single-target pointing clips from
  SGS-HSI with shared actors (\S\ref{sec:corpus}). The two sources provide complementary regimes: spontaneous pointing in dialogue and constrained single-target pointing.

\item A \textbf{task formulation} for referential gesture in continuous
dialogue. 
Systems receive, per test item: audio with a word-level transcript;
a 3D target coordinate in pelvis-centred and world frames; the dominant
pointing hand; a static SMPL-X body shape; and the room's scene graph
(per-object category, bounding-box centre, extent). They must generate
SMPL-X motion containing a pointing gesture directed at the target.
Ground-truth motion, apex frame, referent identity, and evaluator labels
are withheld, so the system decides for itself when the gesture peak
occurs. No oracle apex frame or motion template is provided at inference
time. Phrase intervals are released for training only and are not consumed
by the scorer.

\item An \textbf{evaluation protocol with a ground-truth reference run}. The
  objective half separates \emph{when} a system gestures (temporal alignment,
  Score$_{a}$) from \emph{where} it points (spatial accuracy, Score$_{b}$,
  with tolerance from the annotated referent's extent) and combines them
  multiplicatively (STG, \S\ref{sec:metric}); the subjective half rates rendered
  motion for naturalness in a crowdsourced study (\S\ref{sec:perceptual}).

\item A \textbf{flow-matching baseline}, MM-Conv-Flow: a co-speech
  diffusion-transformer backbone pretrained on BEAT2~\cite{liu2024emage}, adapted to the
  benchmark with target-conditioned residual branches, relational pointing
  losses, and an inference-time directional refinement (\S\ref{sec:baseline-flow}).
  It is trained on the released training split and evaluated under the
  released protocol, so it is a reproducible reference point for future
  systems.
  
\item \textbf{A validity analysis of the protocol.}
The ground-truth reference run exposes complementary limitations of the
objective metrics. RePointer exceeds captured motion on spatial grounding,
showing that evaluator-aligned geometric precision can exceed that exhibited
by human pointing, while its comparison with GT on combined grounding is
sensitive to the temporal scoring parameters. Together with the perceptual
study, these results motivate reporting spatial, temporal, and naturalness
measures separately rather than relying on a single ranking score
(\S\ref{sec:validity}).
\end{enumerate}

\section{Related Work}
The most relevant prior work comprises spatially grounded gesture generation models and datasets that pair communicative motion with explicit scene referents.
\label{sec:related_work}
\subsection{Spatially Grounded Gesture Generation}
Data-driven co-speech gesture generation has progressed from deterministic regression to probabilistic generative sequence models, with 
diffusion~\cite{ao2023gesturediffuclip,deichler2023diffusion} and
flow-matching~\cite{liu2025gesturelsm} models now producing convincing co-speech
motion. Tracking this progress has been supported by the GENEA
Challenges~\cite{kucherenko2023genea}, which fixed a shared corpus and a
large-scale crowdsourced protocol for human-likeness and speech
appropriateness, and by objective measures such as Fr\'echet Gesture
Distance~\cite{yoon2020fgd} and beat alignment that score the motion
distribution and its relation to audio. These models primarily focus on modelling prosodic and semantic alignment rather than
grounding gestures in the surrounding scene. Human--scene and human--object
motion generation instead grounds motion through contact and
affordance~\cite{tokenhsi,tridi,physic,force,wang2024afford,ghosh2026scemos},
but does not address communicative reference.

Work combining these requirements remains limited. A physics-based pointing
agent jointly optimises motion imitation and referential
accuracy~\cite{deichler2023pointing}; subsequent work fine-tunes
OmniControl~\cite{xieomnicontrol} for spatially conditioned
pointing~\cite{deichler2025grounded}; and SIG-Chat~\cite{huang2025sig}
conditions conversational gesture generation on audio, text and 3D target
trajectories. These approaches use different datasets and evaluation
protocols, preventing direct comparison. Our benchmark provides a common
corpus, task and evaluation protocol.

\subsection{Referential Gesture Datasets}
Evaluating referential gesture requires motion paired with scene geometry and
an intended referent. General-purpose embodied datasets provide related
signals: Nymeria~\cite{ma2024nymeria} combines full-body motion with
egocentric video, gaze and 3D geometry, while
Embody3D~\cite{mclean2025embody} additionally captures dialogue.
Synthetic datasets such as ScanERU~\cite{scaneru} and
Ges3ViG~\cite{ges3vig} provide explicit pointing and referent geometry, but
construct pointing instances rather than capture their natural occurrence in
conversation.

Our benchmark builds on MM-Conv~\cite{deichler2026mmconv}: 6.7\,h of
dyadic VR dialogue in five AI2-THOR rooms, in which an instruction-giver
describes the scene to a partner across short role-play scenarios while
full-body motion, speech with word-level alignment, gaze, and the
ground-truth scene graph are captured. Each of its 4,211 referring
expressions is manually linked to its referent object; nearly half (49\%)
are pronominal (``it'', ``that one''), so in many cases non-verbal and contextual cues are needed to resolve the referent.

\section{Corpus and Annotations}
\label{sec:corpus}

\begin{figure*}[t]
  \centering
  \includegraphics[width=\textwidth]{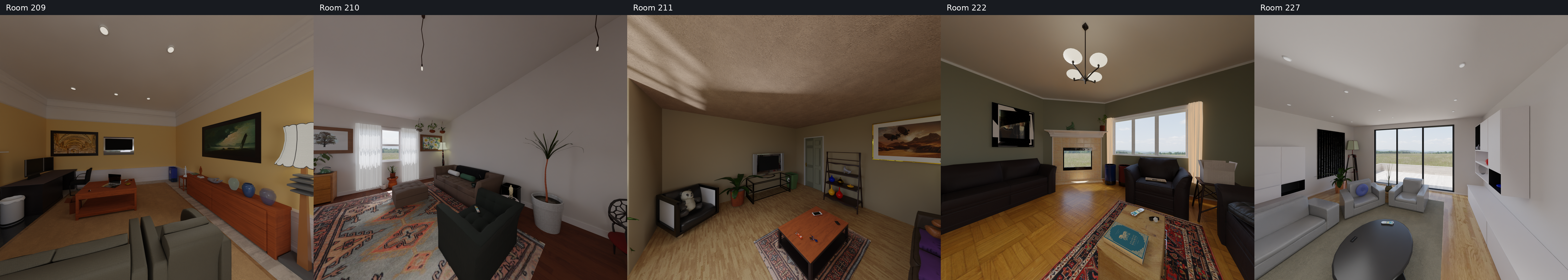}
  \caption{The five AI2-THOR rooms of MM-Conv. Room 211 is held out for evaluation.}
  \label{fig:rooms}
\end{figure*}
\emph{MM-Conv}~\cite{deichler2026mmconv} is the primary data source: 6.7\,h of dyadic VR interaction across five AI2-THOR~\cite{ai2thor} rooms (Fig.~\ref{fig:rooms}), with SMPL-X motion capture at 30\,fps, word-level transcripts, and per-room scene graphs (42--61 objects). Each referring expression is labelled with its ground-truth referent. Capture is dyadic: the addressee's pose is also recorded throughout. We add and release a per-reference pointing annotation layer.
A random-forest classifier using arm-kinematic, scene-context and
speech-timing features guided manual review; human-assigned gesture
tiers determined inclusion. The released benchmark contains 1{,}842 MM-Conv pointing clips
with dominant-hand annotations, divided into training and test
splits as shown in Table~\ref{tab:splits}.  Reference apex frames were estimated
automatically from motion kinematics; manual review established
gesture presence rather than apex timing. This motivates evaluation
in continuous dialogue rather than assuming a gesture at every reference.

\begin{figure*}[t]
  \centering
  \begin{tabular}{@{}c@{\hspace{4pt}}c@{\hspace{4pt}}c@{\hspace{4pt}}c@{}}
    \includegraphics[width=0.245\textwidth]{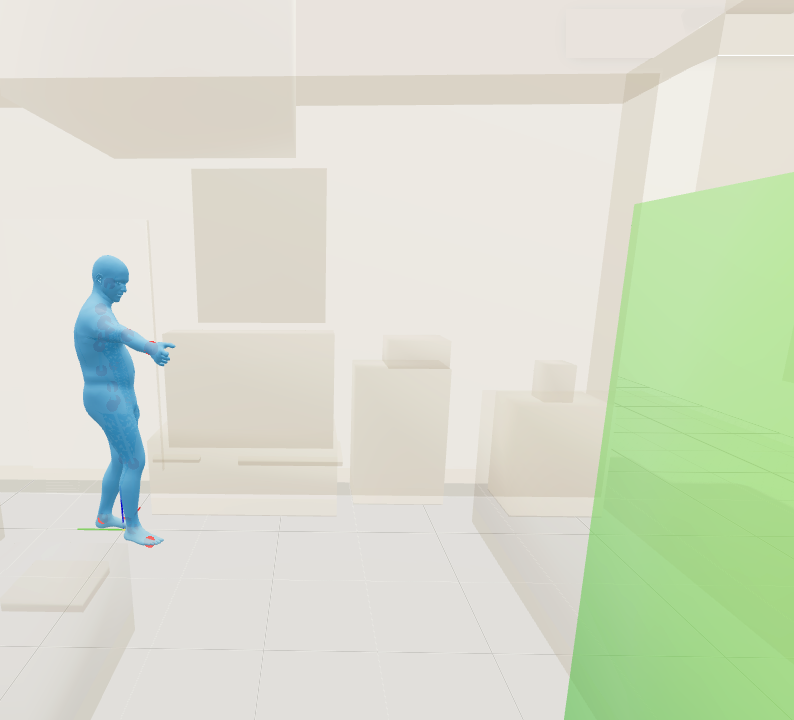} &
    \includegraphics[width=0.245\textwidth]{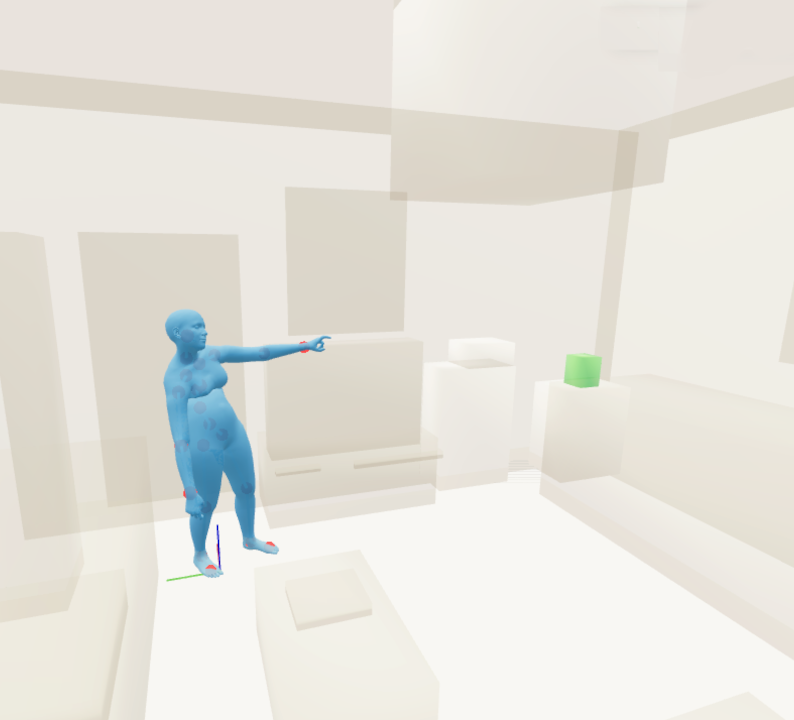} &
    \includegraphics[width=0.245\textwidth]{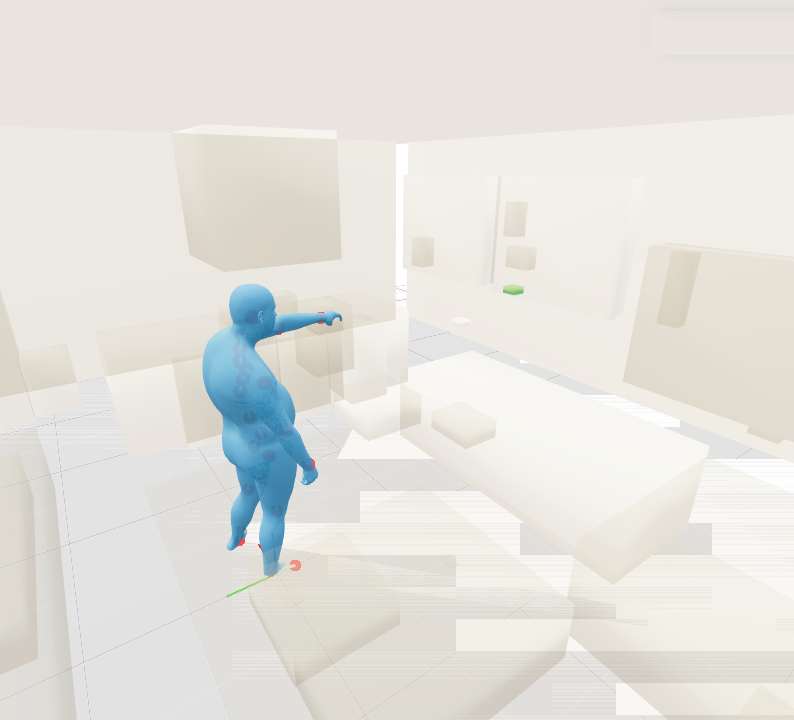} &
    \includegraphics[width=0.245\textwidth]{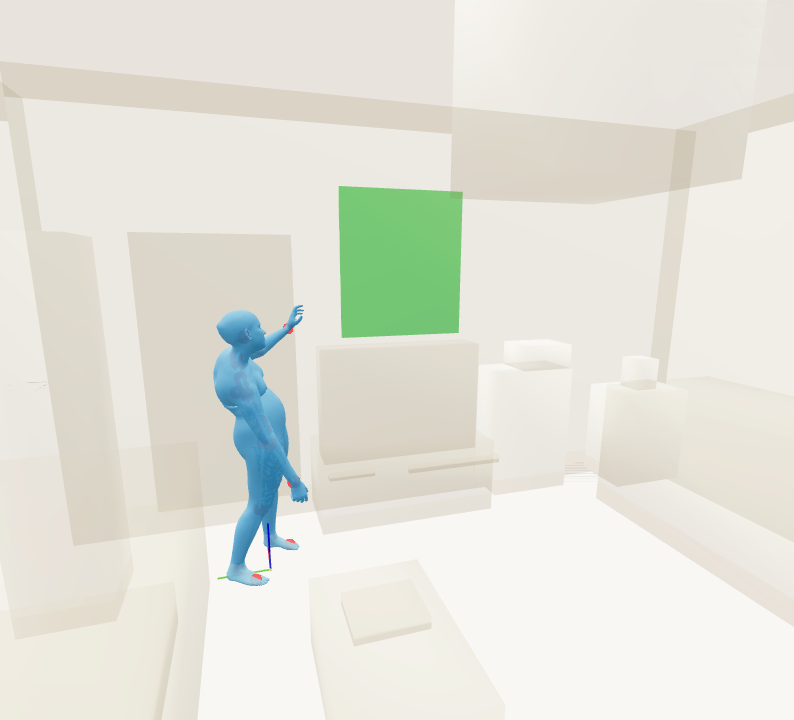}  \\
    {\small (a)} &
    {\small (b) } &
    {\small (c)} &
    {\small (d) }
  \end{tabular}
  \caption{Training data samples across target configurations.
    The blue mesh shows the captured SMPL-X body, and the green
    volume marks the target object. Visualisation tool based on  ~\cite{yi2025viser}, released with the benchmark.}
  \label{fig:qualitative}
\end{figure*}

\emph{SGS-HSI}~\cite{deichler2022towards} adds 1{,}138 isolated single-target
pointing clips: clean geometry, one target coordinate and apex frame per clip,
no scene graph. Motion is real capture, augmented
and retargeted to the MM-Conv actors. Speech is synthesised and matched to the gestures, so phrase–gesture alignment is constructed rather than naturally occurring. Figure~\ref{fig:qualitative} illustrates representative training examples
and the range of target configurations.

\paragraph*{Splits.} One room is held out (Table~\ref{tab:splits}); the fixed 100-item evaluation manifest is randomly sampled from the held-out MM-Conv clips, stratified by speaker to maintain approximately balanced representation across the five actors. SGS-HSI is split by source gesture, so no test clip shares a source with training. Phrase intervals are released for training only and never consumed by the scorer.

\begin{figure}[tb]
  \centering
  \includegraphics[width=\linewidth]{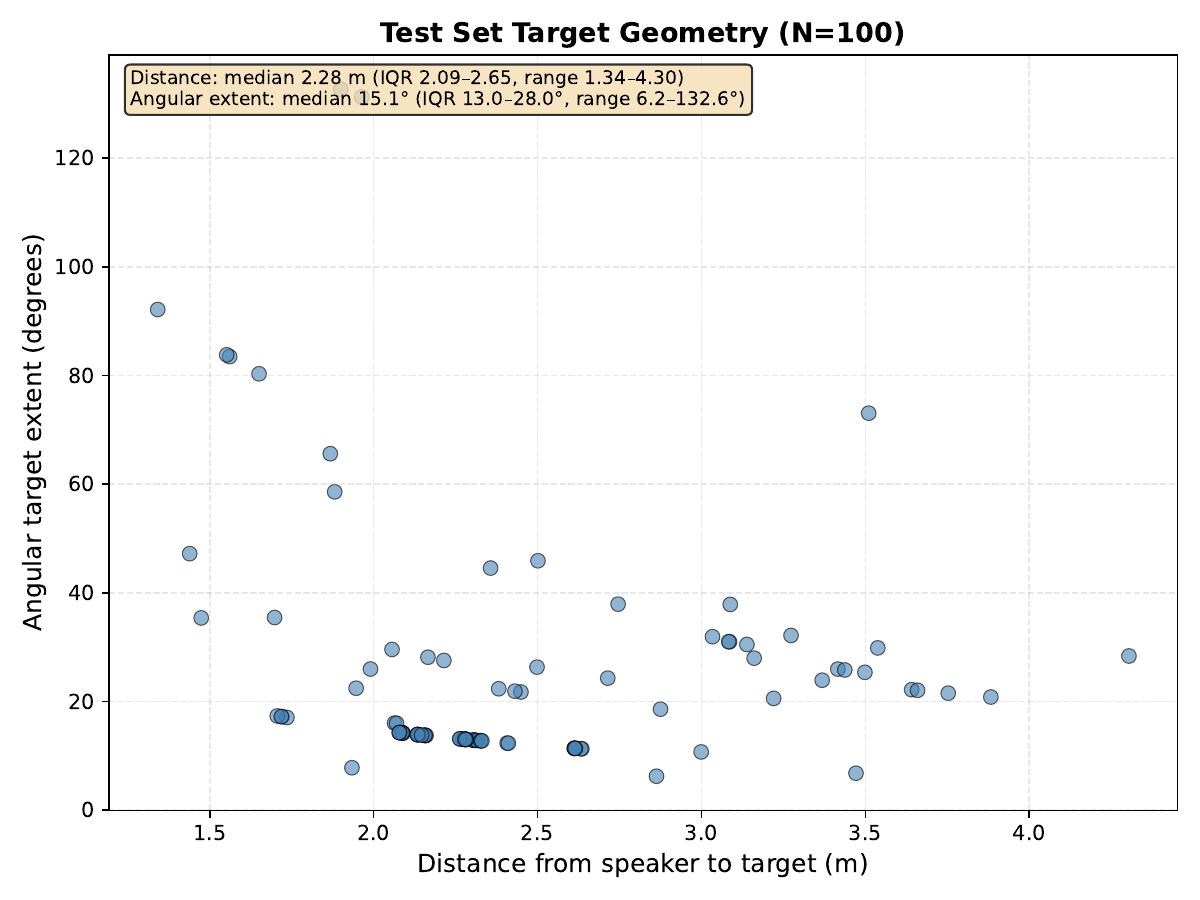}
  \caption{Geometry of the 100 held-out test targets. Each point represents
  one test clip. Targets span substantial variation in speaker--target
  distance and apparent angular extent.}
  \label{fig:target-geometry}
\end{figure}

\paragraph*{Test-set geometry.}
The fixed 100-item evaluation set spans 75 unique target instances across
21 object categories, with no individual target appearing in more than five
clips. Targets lie 1.3--4.3\,m from the speaker, vary widely in apparent
size (Fig.~\ref{fig:target-geometry}), and cover nearly the full azimuth
range ($-176$\textdegree{} to $177$\textdegree{}) and elevations from
$-32$\textdegree{} to $55$\textdegree{}, so the set includes targets behind
the speaker and both above and below shoulder height.

\paragraph*{Release.} SMPL-X motion and audio at 30\,fps, transcripts, scene-graph JSON, and a sample-level annotation table are available at \url{https://huggingface.co/datasets/hsi-workshop/referential-gesture-challenge}.

\begin{table}[tb]
\caption{Scene-disjoint split statistics on MM-Conv and SGS-HSI. Final benchmark scores (Table~\ref{tab:objective}) use a fixed 100-item evaluation set drawn from the held-out MM-Conv clips.}
  \label{tab:splits}
  \centering
  \begin{tabular}{@{}lcccc@{}}
    \toprule
    Split & Rooms & MM-Conv & SGS-HSI & Total \\
    \midrule
    Train & 4 rooms & 1,503 & 1,022 & 2,525 \\
    Test  & 1 room  &   339 &   116 &   455 \\
    \bottomrule
  \end{tabular}
\end{table}


\section{Benchmark}
\label{sec:benchmark}

\subsection{Task Formulation}
\label{sec:task}

Systems receive, per test item: audio with a word-level transcript; a 3D target
coordinate in pelvis-centred and world frames; the dominant pointing hand; a
static SMPL-X body shape; and the room's scene graph (per-object category,
bounding-box centre, extent). They must generate SMPL-X motion containing a
pointing gesture directed at the target. Ground-truth motion, apex frame,
referent identity, and evaluator labels are withheld, so the system decides for
itself when the gesture peak occurs. Since the target coordinate is supplied,
the scene graph is optional. It can be used for reasoning about distractors or
occlusion, or ignored. Supplying
the coordinate isolates gesture generation from referent resolution; the harder
variant that withholds it will be the focus of future benchmark releases.

\subsection{Evaluation Protocol}
\label{sec:metric}
We evaluate two complementary aspects of generated gestures: motion quality
and referential grounding. Standard motion-quality metrics characterise the
motion distribution, dynamics, and audio--motion relation, but do not measure
whether a gesture successfully refers to the intended object. 
We therefore evaluate referential grounding through two components: temporal alignment (Score$_a$), measuring the timing of the detected gesture apex relative to the reference apex, and spatial grounding (Score$_b$), measuring whether the gesture indicates the intended 3D target. Their product gives the combined spatio-temporal grounding score (STG). All grounding scores are computed automatically from generated SMPL-X
motion and are bounded in $[0,1]$; systems provide no apex frame or
other evaluation metadata.

\paragraph*{Temporal alignment (Score$_a$).}
\label{sec:apex-detector}
A frozen kinematic apex detector, implemented as logistic regression over
pooled per-frame body features and trained on recordings disjoint from the
test set, locates the pointing apex $\hat a$ in each generated motion.
During training, frames within $\pm3$ frames of the reference apex are
labelled positive; this supervision window is distinct from the scoring
tolerance below.
We compare the detected apex with the reference apex $a_{\mathrm{ref}}$ using
\begin{gather*}
e_t = |\hat a-a_{\mathrm{ref}}|,\qquad d_t = \max(0,\,e_t-\tau),\\
\mathrm{Score}_a = \exp\!\left(-\frac{d_t^2}{2\sigma_t^2}\right).
\end{gather*}
We use the same tolerance $\tau=10$ frames (0.33\,s) and decay
$\sigma_t=15$ frames (0.5\,s) for every clip. The fixed tolerance avoids
making temporal credit depend on reference-gesture duration. These are
benchmark design parameters rather than empirically estimated human timing
thresholds; sensitivity analysis in Appendix~\ref{sec:supp-tau} shows that
the RePointer--MM-Conv-Flow comparison is robust to their choice.
Because the reference apex is itself estimated from motion kinematics, and the detector is trained on kinematic features of captured motion, the GT timing score reflects agreement between two kinematic definitions of the apex; it is a diagnostic, not independent validation of detector reliability.

\paragraph*{Spatial accuracy (Score$_{b}$).}
At the apex $\hat{a}$ detected by Score$_{a}$, we cast a semi-infinite ray from the wrist along the elbow-to-wrist direction of the annotated pointing hand. This forearm direction provides a simple, reproducible proxy for arm-based pointing, without modelling finger articulation or gaze. Let $d_{\text{miss}}$ be the ray's distance to the target's axis-aligned bounding box (zero on intersection) and $d_{\text{centroid}}$ be the wrist-to-centroid
distance. The extent-aware angular error
$\theta_{\text{eff}}=\arctan(d_{\text{miss}}/d_{\text{centroid}})$ is mapped to
$\text{Score}_{b}=\exp(-\theta_{\text{eff}}^{2}/2\sigma^{2})$ with a fixed decay scale $\sigma=20^{\circ}$. Target extent is accounted for directly through \(d_{\mathrm{miss}}\): a ray intersecting the target bounding box has zero effective error. Thus, $\sigma$ controls how rapidly the score decays once the pointing ray misses the target rather than defining a tolerance for target size. As with the temporal parameters, $\sigma$ is a benchmark design parameter rather than an empirically estimated perceptual threshold.

\paragraph*{Combined grounding (STG).}
STG $=$ Score$_{a}\times$Score$_{b}$. The product penalises errors in either temporal or spatial grounding, while reporting the two factors separately exposes the source of a low combined score.

\paragraph*{Motion quality.}
We report motion quality separately from grounding using EMAGE~\cite{liu2024emage}
Fr\'echet Gesture Distance (FGD), average jerk, L1 velocity diversity,
and beat alignment. These metrics characterise complementary distributional,
kinematic, and audio--motion properties of generated motion; they do not
measure whether a gesture successfully refers to the intended target.
FGD is computed against the training-reference feature distribution.


\subsection{MM-Conv-Flow}
\label{sec:baseline-flow}
The baseline is a reference point, developed for this benchmark. It is trained on
the benchmark training split and evaluated on the held-out test split, so its
scores are directly comparable to any system evaluated under the protocol. Full hyperparameters are in Appendix~\ref{sec:supp-mmconvflow}.

\subsubsection*{Co-Speech Gesture Backbone (Phase~1)}
\label{sec:p1}
The backbone is a 170.5\,M-parameter, 12-layer Diffusion Transformer (MotionDiT)
trained on BEAT2~\cite{liu2024emage} via conditional flow matching. With $x_0 \sim \mathcal{N}(\mathbf{0}, \mathbf{I})$ the noise sample and $x_1$ the ground-truth motion, we use the linear optimal-transport interpolant
\begin{equation}
  \vx_t = t\,\vx_1 + (1-t)\,\vx_0,\qquad t\in[0,1],
  \label{eq:ot-interpolant}
\end{equation}
whose conditional velocity field is constant along the path,
\begin{equation}
  v_{\text{target}}(x_t, t) = \frac{\mathrm{d}x_t}{\mathrm{d}t} = x_1 - x_0,
  \label{eq:velocity}
\end{equation}

so the network $\vv_\theta$ is trained with the regression objective
\begin{equation}
  \mathcal{L}_{\text{CFM}}
  = \mathbb{E}_{t,\vx_0,\vx_1,\vc}
    \bigl\|\vv_\theta(\vx_t,t,\vc)-(\vx_1-\vx_0)\bigr\|_2^2 .
  \label{eq:cfm-loss}
\end{equation}
Motion is a 300-dimensional SMPL-X representation
($25\text{ joints}\times[\,3\text{ positions},\,6\text{D rotations},\,3\text{ velocities}\,]$)
at 30\,fps over $T{=}180$ frames (6\,s). Speech conditioning $\vc$ uses a frozen
306.8\,M-parameter CSMP encoder~\cite{deichler2023diffusion} over aligned audio
and text data2vec features, which provides (i)~a pooled global
embedding modulating all 12 layers via adaptive layer normalisation with
progressive scaling, and (ii)~frame-level features attended to by layers
4--11 through windowed cross-attention (${\pm}3$ frames, ${\approx}{\pm}100$\,ms).
Two mechanisms carry audio--gesture synchrony: an adaptive temporal attention that
learns per-sample importance weights for beat-relevant audio frames, added as a
learned bias to the cross-attention scores before the softmax; and the windowed
mask itself, which restricts each motion frame to local audio context and
prevents long-range semantic leakage. Auxiliary losses enforce temporal smoothness (total variation, velocity, jerk, jerk-distribution matching), all weighted by $t^2$; a bone-length loss is applied unweighted.
Phase~1 is trained for 1{,}100 epochs on a single RTX~3090, reaching
FGD~$<1.0$ on BEAT2 validation. Sampling uses 50 Euler steps.

  \subsubsection*{Spatial Conditioning (Phase~2)}
  \label{sec:p2}
  Phase~2 injects target geometry into the Phase~1 backbone while preserving
  co-speech quality.  The 3D pelvis-relative target $\mathbf{t}$ is encoded
  as

  \begin{equation}
\vc_{\text{spatial}}
  = \mathrm{MLP}\!\left(\left[\tfrac{\mathbf{t}}{\|\mathbf{t}\|},\;
    \log(1{+}\|\mathbf{t}\|)\right]\right)
  + \mathbf{e}_{\text{hand}},
\label{eq:dirdist}
\end{equation}
factoring unit direction and log-distance through a small MLP ($4\to256\to256\to768$)
with a learned hand embedding, giving $\vc_{\text{spatial}}\in\mathbb{R}^{768}$.

Both variants below (Modes~A and~B) load the Phase~1 parameters and introduce functionally identity-initialised spatial residual branches: the static-bias projections and spatial cross-attention output projections are zero-initialised, while the target encoder uses small nonzero initialisation to preserve gradient flow:
  
  \emph{Mode~A (frozen adapter).} 
  The backbone is frozen; layers 4--11
  receive a gated residual bias:
  \begin{equation}
  \vx \leftarrow \vx + \sigma(g_\ell)\,W_\ell\,\vc_{\text{spatial}},
  \label{eq:modea}
  \end{equation}
  with $g_\ell$ initialised to $-2.0$ and $W_\ell$ zero-initialised.  This tests
  whether pointing can be added without modifying the pretrained motion prior.

  \emph{Mode~B (unfrozen + cross-attention).}  The backbone is unfrozen, retains
  the static bias of Eq.~\eqref{eq:modea} on layers 4--11, and layers 8--11
  additionally attend to $\vc_{\text{spatial}}$ via
  single-token cross-attention:
  \begin{equation}
  \vx \leftarrow \vx + \sigma(g_\ell^{\text{xattn}})\,r_t\,
    \mathrm{CrossAttn}(\mathrm{LN}(\vx),\;\vc_{\text{spatial}}),
  \label{eq:modeb}
  \end{equation}
  where $r_t = \sigma(\mathrm{LN}(\vx) \cdot \vc_{\text{spatial}}/\sqrt{d})$ is a
  per-frame routing gate based on query-key compatibility, $g_\ell^{\text{xattn}}$
  is initialised to $-3.0$, and output projections are zero-initialised.
  Mode~B has higher capacity but risks degrading the co-speech prior if spatial
  gradients dominate.

\paragraph*{Relational pointing losses.}
Spatial supervision uses forearm--target alignment and arm extension,
applied only to samples with targets:
\begin{align}
\mathcal{L}_{\text{align}}
  &= 1 - \cos(\mathbf{d}_{\text{forearm}},\;\mathbf{d}_{\text{target}}),
  \label{eq:flow-align}\\
\mathcal{L}_{\text{extend}}
  &= \max(0,\; r_{\text{rest}} - r_{\text{current}}),
  \label{eq:flow-extend}
\end{align}
where $\mathbf{d}_{\text{forearm}}$ is the elbow-to-wrist direction,
$\mathbf{d}_{\text{target}}$ the elbow-to-target direction, arm extension $r$ is the pelvis-to-selected-wrist distance, with a hinge penalty below a fixed $0.35\,\mathrm{m}$ threshold. Both terms are computed on every valid frame and aggregated with a soft-min over time (temperature 0.1), so the frames where the arm points best dominate the loss; the arm is therefore not required to aim at the target throughout the clip, and no apex annotation is used. The combined relational loss is weighted by  $t^2$ and evaluated only on target-bearing rows; BEAT2 rows have zero spatial weight and bypass spatial conditioning.

Phase~2 trains for 200 epochs (batch 64, cosine schedule, single RTX~3090) with a
70/30 mix of pointing clips (SGS-HSI + MM-Conv) and BEAT2, audio dropout
$p{=}0.3$, explicit jerk ($w{=}0.05$) and velocity ($w{=}0.01$) smoothness
losses, and layer-wise LR decay (factor 0.75) with a separate higher learning
rate for the spatial parameters.


\subsubsection*{Inference-Time Directional Refinement}
\label{sec:nudge}
Inspired by the predict--correct--recompose structure of ProjFlow~\cite{Watanabe_2026_CVPR}, we introduce a deterministic directional nudge: we form a per-frame wrist-position target from the current clean-endpoint estimate and apply a partial kinematic correction rather than ProjFlow's exact linear projection.

\begin{equation}
    \mathbf{t}_f^{\mathrm{world}}=\mathbf{p}_f + \mathbf{t}^{\mathrm{pelvis}}
\end{equation}
\begin{equation}
\mathbf{w}_{\text{desired,f}}
  = \mathbf{s}_f
  + \frac{\mathbf{t}_f^{world}-\mathbf{s}_f}{\|\mathbf{t}_f^{world}-\mathbf{s}_f\|}
    \cdot \|\mathbf{w}_f-\mathbf{s}_f\|,
\label{eq:desired-wrist}
\end{equation}
with $\mathbf{s}$ the shoulder and $\mathbf{w}$ the current wrist.  The release
supplies the target in both frames; the baseline conditions on the
pelvis-relative one (Eq.~\eqref{eq:dirdist}), so the world-frame target is
reconstructed from the estimated pelvis at each step.  This target preserves the
current arm length $\|\mathbf{w}-\mathbf{s}\|$; the weighted propagation below
preserves it only approximately.  Gating uses the model's own arm extension
rather than frame indices:
\begin{equation}
\alpha_f
  = \underbrace{\frac{\|\mathbf{w}_f-\mathbf{s}_f\|}
                     {\max_{f'}\|\mathbf{w}_{f'}-\mathbf{s}_{f'}\|}}
               _{\text{extension ratio}}
  \cdot \beta
  \cdot \Bigl(\tfrac{i}{N-1}\Bigr)^{\!\gamma}
\label{eq:nudge-weight}
\end{equation}
for frame $f$ at Euler step $i\in\{0,\dots,N-1\}$ of $N$. The refinement parameters $\beta=0.5$ and $\gamma=2$ were selected on a
validation split drawn from the training rooms and fixed before evaluation
on the held-out room.
Each step predicts the clean endpoint $\hat{\vx}_1=\vx_t+(1-t_{\text{next}})\vv$, applies the weighted correction
$\Delta\mathbf{w}_f=\alpha_f(\mathbf{w}_{\text{desired},f}-\mathbf{w}_f)$ to the arm chain (wrist ${\times}1.0$, elbow ${\times}0.5$, shoulder ${\times}0.25$), and reconstructs $\vx_t$ from the corrected endpoint; we omit ProjFlow's stochastic recomposition step. The fixed chain weights approximate the locality of ProjFlow's kinematic correction, with the largest adjustment at the observed joint and decreasing influence along the kinematic tree. While the training loss aligns the elbow--wrist vector, the refinement steers the arm toward the shoulder--target ray.

Mode~B substantially improves spatial grounding over the frozen adapter, while the directional refinement further reduces angular error from 19.5\textdegree{} to 11.0\textdegree{}.

\begin{table}[tb]
\caption{Development-stage evaluation of MM-Conv-Flow variants on the full 455-clip held-out set (MM-Conv + SGS-HSI), using an earlier hold-based temporal scorer. Included to show the relative effect of the baseline design choices; temporal and STG values are not directly comparable to the final 100-item MM-Conv evaluation in Table~\ref{tab:objective}. \(^{*}\)Configuration used in the final evaluation.}
\label{tab:baseline-ablation}
\centering
\small
\begin{tabular}{@{}lcccc@{}}
\toprule
Variant & Score$_{a}$ $\uparrow$ & Score$_{b}$ $\uparrow$ & STG $\uparrow$
        & $\theta_{\text{eff}}$ $\downarrow$ 
         \\
\midrule
GT MoCap             & 0.909 & 0.765 & 0.718 & 11.8\textdegree   \\
\midrule
Mode A (frozen)      & 0.527 & 0.315 & 0.173 & 36.2\textdegree  \\
Mode B (unfrozen)    & 0.593 & 0.593 & 0.374 & 19.5\textdegree   \\
Mode B + nudge$^{*}$ & 0.581 & 0.743 & 0.463 & 11.0\textdegree  \\
\bottomrule
\end{tabular}
\end{table}

\section{Evaluation}
We evaluate the baseline, one independently developed system, and captured motion.

\subsection{RePointer}
\label{sec:repointer}

RePointer~\cite{he2026repointer} is a spatially conditioned retrieve--generate--score--select system. It first
retrieves a full-body SMPL-X anchor from a memory of natural pointing clips,
indexed by target geometry, pointing hand, predicted gesture timing, and apex
ratio, with donors from the held-out fold excluded. Realisation operators then
adapt the anchor: an apex-preserving temporal warp that resamples pre- and
post-apex frames separately to the query timing; a body-shape-aware target-ray
inverse kinematics step minimising
\begin{equation}
\mathcal{L}_{\mathrm{IK}}= 1-\widehat{(\mathbf{w}-\mathbf{e})}^{\top}
\widehat{(\mathbf{t}-\mathbf{e})} +\lambda_p\|\Delta\boldsymbol{\theta}\|_2^2
+\lambda_s\mathcal{L}_{\mathrm{smooth}}
\label{eq:repointer-ik}
\end{equation}
over the collar--shoulder--elbow--wrist chain; and a short tapered hold producing
one high-extension, low-velocity pointing region. Root translation, lower body,
and the non-active arm are preserved from the anchor unchanged, so aiming is
achieved by arm articulation alone. Cross-fitted scorers rank the anchor against
generated hypotheses using only public test-time inputs and motion-derived
features, and a selector returns one candidate unchanged. No test ground-truth
motion, apex, object identity, or evaluator labels are used. Eq.~\eqref{eq:repointer-ik} makes forearm--target alignment an explicit inference-time
objective---the same quantity Score$_{b}$ reads.

\subsection{Objective Evaluation}

\begin{table}[tb]
\caption{Objective evaluation on 100 held-out MM-Conv clips.
Grounding scores are the mean $\pm$ SD across clips.}
\label{tab:objective}
\centering
\small
\setlength{\tabcolsep}{4pt}
\begin{tabular}{@{}lcccc@{}}
\toprule
Metric & GT & MM-Conv-Flow & RePointer & Dir. \\
\midrule
\multicolumn{5}{@{}l}{\emph{Spatial}} \\
\quad Score$_b$
 & $0.767{\pm}0.223$ & $0.691{\pm}0.291$ & $0.905{\pm}0.195$ & $\uparrow$ \\

\multicolumn{5}{@{}l}{\emph{Temporal}} \\
\quad Score$_a$
 & $0.941{\pm}0.210$ & $0.499{\pm}0.402$ & $0.784{\pm}0.278$ & $\uparrow$ \\
\multicolumn{5}{@{}l}{\emph{Spatio-Temporal}} \\
\quad STG
 & $0.720{\pm}0.265$ & $0.361{\pm}0.365$ & $0.719{\pm}0.307$ & $\uparrow$ \\

\multicolumn{5}{@{}l}{\emph{Motion Quality}} \\
\quad EMAGE FGD
 & 4.738 & 4.551 & 2.364 & $\downarrow$ \\
\quad Avg. Jerk
 & 0.261 & 0.687 & 0.543 & $\downarrow$ \\
\quad L1 Vel. Div.
 & 0.246 & 0.178 & 0.760 & --- \\
\quad Beat Align.
 & 0.318 & 0.272 & 0.380 & $\uparrow$ \\
\bottomrule
\end{tabular}

\vspace{0.3em}
\parbox{\linewidth}{\footnotesize
STG is the per-clip product Score$_a\times$Score$_b$, then averaged.
EMAGE FGD is computed against the training-reference distribution.}
\end{table}

\subsubsection{Grounding Analysis}
\label{sec:objective-results}

\paragraph*{Spatial Grounding.}
RePointer achieves the highest spatial grounding score
(Score$_b{=}0.905$), compared with 0.691 for MM-Conv-Flow and 0.767 for
GT. Its advantage over MM-Conv-Flow is $+0.214$ [95\% CI:
$+0.163$, $+0.272$], using a bootstrap clustered over the 59 source
recordings. RePointer also scores above GT ($\Delta{=}{+}0.140$
[95\% CI: $+0.088$, $+0.196$]). Because RePointer explicitly optimises forearm--target geometry closely related to Score$_b$, this result shows that the metric rewards
evaluator-aligned geometric precision beyond that exhibited by the
captured motion. Score$_b$ alone cannot distinguish such precision from
communicative accuracy: whether it helps an observer identify the
intended referent is a separate question (§\ref{sec:validity}).

\begin{figure}[tb]
    \centering
    \includegraphics[width=0.8\columnwidth]{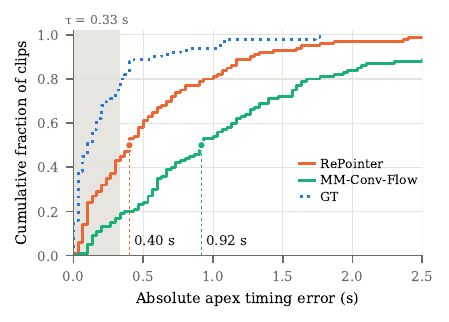}
    \caption{Temporal alignment in seconds. Cumulative distribution of
    absolute apex timing error over the 100 test clips. The shaded region
    marks the primary tolerance $\tau=0.33$\,s; markers indicate median
    errors of 0.40\,s for RePointer and 0.92\,s for MM-Conv-Flow.
    For GT, the same frozen apex detector is applied to the captured
    reference motion and is shown as a detector diagnostic rather than an
    estimate of human timing variability.}
    \label{fig:temporal-error}
\end{figure}

\paragraph*{Temporal Alignment.}
RePointer is substantially better temporally aligned than MM-Conv-Flow,
achieving Score$_a{=}0.784$ versus 0.499. In raw timing, the median absolute
apex errors are 0.40\,s and 0.92\,s, respectively
(Fig.~\ref{fig:temporal-error}). In MM-Conv-Flow, temporal placement is left to the speech-conditioned
generative backbone; the larger timing error shows that this model
does not consistently place the referential apex precisely under
the benchmark's detector-based criterion. This advantage is robust to the temporal
metric design: RePointer scores higher across all 806 tested combinations
of $\tau$ and $\sigma_t$ (Appendix~\ref{sec:supp-tau}). GT obtains
Score$_a{=}0.941$; because the reference apex is kinematically defined and
the same detector is applied to captured motion, we treat this value as a
detector/reference diagnostic rather than a human timing baseline.

\paragraph*{Combined Spatio-Temporal Grounding.}
STG is the per-clip product Score$_a\times$Score$_b$, averaged over clips.
Under the primary temporal parameters, RePointer and GT obtain nearly
identical mean STG (0.719 and 0.720, respectively), while MM-Conv-Flow
obtains 0.361. RePointer therefore substantially exceeds MM-Conv-Flow on
the combined measure. In contrast, the RePointer--GT ordering changes with
the temporal tolerance and decay parameters (Appendix~\ref{sec:supp-tau}).
We therefore draw no conclusion about RePointer relative to captured motion
from STG and report the temporal and spatial factors separately.

\subsubsection{Motion Quality.}
RePointer obtains the lowest EMAGE FGD (2.364), compared with MM-Conv-Flow (4.551) and held-out GT (4.738), using the same training reference. This advantage should be interpreted in light of its retrieval-based construction: RePointer initialises from human motion in the retrieval base, which overlaps the FGD reference distribution, and preserves the root, lower body, and non-active arm.

Its average jerk (0.543) lies between MM-Conv-Flow (0.687) and GT (0.261). RePointer has substantially higher L1 velocity diversity than both MM-Conv-Flow and GT (0.760 vs. 0.178 and 0.246, respectively), while its beat alignment is moderately higher than both (0.380 vs. 0.272 and 0.318). Despite these differences, RePointer is not rated more natural than MM-Conv-Flow (MOS 2.95 vs. 3.11, \(p=.071\); \S\ref{sec:perceptual}). Thus, its lower FGD and higher diversity and beat-alignment scores do not establish greater perceived naturalness; the metrics capture different properties of motion.


\subsection{Perceptual Evaluation}
\label{sec:perceptual}
Objective grounding scores do not capture whether generated motion looks
human-like. We ran a crowdsourced MOS study to complement the spatio-temporal
metrics.

\subsubsection{Protocol.}
We sampled 23 referring-expression clips from the held-out test room and
rendered each under three conditions: ground-truth motion capture (GT),
MM-Conv-Flow, and RePointer. All conditions use the same speech audio, avatar, room, and camera;
only body motion varies. Conditions were counterbalanced with a cyclic Latin
square: participants in rotation~$g$ saw clip~$i$ in condition~$c_{(i+g)\bmod
3}$, so each participant rated every clip exactly once and, across three
rotations, every clip$\times$condition cell was covered. Trial order was
randomised with no more than two consecutive trials from the same condition.

Participants were recruited on Prolific (English first language, desktop) and
rated each clip on a five-point scale (1\,=\,very unnatural, 5\,=\,very
natural) after watching it in full; playback controls were disabled, and the
rating was gated on ${\geq}\,85\%$ of the clip having been played. Each session
contained 23 stimulus trials plus two attention checks. Sessions took a median
of 5.5~minutes and were paid \pounds1.25.
\subsubsection{Results.}
\begin{figure*}[tb]
  \centering
  \includegraphics[width=\textwidth]{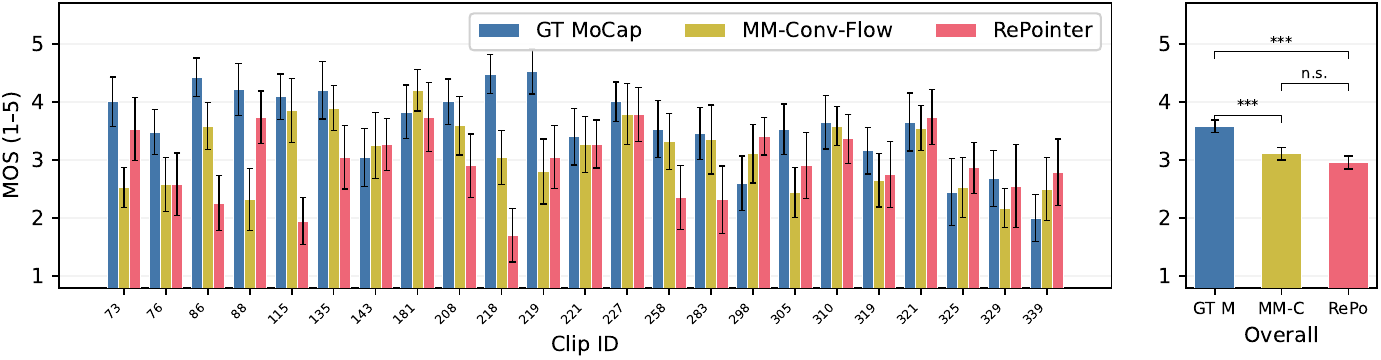}
  \caption{Perceptual evaluation of motion naturalness.
  \emph{Left}: mean opinion scores (MOS; 1--5) for each of the 23
  evaluated clips under GT MoCap, MM-Conv-Flow, and RePointer.
  \emph{Right}: overall means. Error bars show 95\% CIs of the mean.
  Significance is computed using Wilcoxon signed-rank tests on
  per-participant condition means with Holm correction
  ($^{***}p<.001$; n.s.: $p\geq.05$).}
  \label{fig:mos}
\end{figure*}

Sixty-two participants, all of whom passed both attention checks,
contributed 1,410 ratings (467--473 per condition;
Table~\ref{tab:mos}; Fig.~\ref{fig:mos}); 16 trials were excluded for video load error. GT was rated significantly
more natural than both MM-Conv-Flow ($W=222$,
$p_{\mathrm{Holm}}<10^{-6}$) and RePointer ($W=76$,
$p_{\mathrm{Holm}}<10^{-9}$). MM-Conv-Flow received a numerically
higher MOS than RePointer (3.11 versus 2.95), but the difference was
not statistically significant ($W=694$, $p_{\mathrm{Holm}}=.071$).

\begin{table}[tb]
\caption{Mean opinion scores for motion naturalness (1--5). CIs are 95\%
CIs of the mean rating; $n$ denotes individual ratings. Pairwise tests use
Wilcoxon signed-rank tests on per-participant condition means with Holm
correction ($N=62$; one zero difference discarded in the MM-Conv-Flow vs.\
RePointer test, $N=61$; \textsuperscript{***}$p<.001$; n.s.: $p\geq.05$).}

\label{tab:mos}
\centering
\begin{tabular}{@{}lcccc@{}}
  \toprule
  Condition & MOS & 95\% CI & $n$ & vs.\ GT \\
  \midrule
  GT MoCap      & 3.58 & $\pm$\,0.10 & 473 & --- \\
  MM-Conv-Flow  & 3.11 & $\pm$\,0.10 & 470 & *** \\
  RePointer     & 2.95 & $\pm$\,0.11 & 467 & *** \\
  \midrule
  \multicolumn{4}{@{}l}{MM-Conv-Flow vs.\ RePointer} & n.s. \\
  \bottomrule
\end{tabular}
\end{table}

\section{Validity Analysis}
\label{sec:validity}

The benchmark exposes two limitations that are important when interpreting
referential-gesture metrics.

\paragraph*{Geometric grounding is not a proxy for perceived naturalness.}
RePointer achieves Score$_b{=}0.905$, above captured motion (0.767), by
explicitly optimising forearm--target geometry at inference time.
Score$_b$ therefore rewards geometric precision beyond that exhibited by
the captured gestures, but it does not establish that this precision is
communicatively useful: human pointing is not aimed along the
elbow--wrist ray, and observers may resolve referents from coarser cues.
At the same time, RePointer is not rated more natural than MM-Conv-Flow,
so its geometric advantage does not transfer to perceived motion quality.
Geometric grounding and human-like motion quality must therefore be
evaluated separately, and neither substitutes for a direct test of
communicative success.

A qualitative failure mode of the retrieval-based system is that the retrieved motion prior can carry gesture semantics unrelated to the current utterance. For example, we observed a case (Appendix~\ref{sec:supp-video}, clip 298) in which the retrieved anchor contained an iconic object-interaction-like motion from the training corpus, while the subsequent geometric adaptation redirected the arm toward the correct referent. Such cases can obtain good spatial grounding despite semantically inappropriate residual motion, illustrating that Score$_b$ evaluates target-directed geometry rather than the semantic appropriateness of the complete gesture.

\paragraph*{Combined grounding depends on temporal metric design.}
Under the primary temporal parameters, RePointer and GT have nearly identical
STG (0.719 and 0.720), but their ordering changes as the temporal tolerance
and decay are varied (Appendix~\ref{sec:supp-tau}). In contrast, RePointer remains
above MM-Conv-Flow across all tested temporal parameterisations. Thus the
system comparison is robust, whereas an aggregate RePointer--GT comparison
is not. Reporting Score$_a$ and Score$_b$ separately, together with raw
timing error, makes the source of an STG result explicit.

\section{Limitations}
\emph{Data.} The benchmark rests on five rooms of one VR corpus, with a single held-out room supplying all 100 test clips. SGS-HSI extends the dataset with clean single-target pointing, but its speech is synthesised and phrase--gesture alignment is constructed, so it does not extend the naturalistic referential domain. Larger referential corpora spanning more scenes (beyond indoor household rooms) and more speakers would benefit the benchmark.

\emph{Task.} Supplying the target isolates gesture generation from reference resolution. Future variants could require joint generation of language, speech, and motion as a benchmark task for referential communication in embodied agents.

\emph{Protocol.} Score$_b$ measures target-directed forearm geometry.
The retrieval-based system shows that this quantity can be optimised
directly at inference time, so Score$_b$ should be read as a geometric
precision measure, not as evidence of communicative success.
The spatial score also depends on estimated elbow and wrist positions;
sensitivity to pose-estimation or calibration errors remains unquantified.
Score$_a$ likewise depends on a frozen apex detector and a fixed tolerance
that are benchmark design choices rather than perceptual thresholds.

\emph{Perceptual evaluation.} The MOS study measures naturalness only. Whether observers can identify the intended referent from generated motion---the communicative question the benchmark ultimately targets---is not tested here. The current test set is not suited to such a study: items were sampled for speaker balance and target diversity, not for controlled distractor structure, so many targets have no comparably placed alternatives and a forced-choice task would be trivial on some items and ill-posed on others. A referent identification study requires a dedicated item set in which each target has matched distractors at controlled angular separations, so that chance level and difficulty are known. We consider this a necessary complement to Score$_b$ and plan it as the next benchmark release.

\section{Discussion}

\paragraph*{Neither system learns end-to-end spatial grounding.}
Both evaluated systems use the supplied target coordinate to improve spatial grounding at inference time. MM-Conv-Flow's directional refinement
(\S\ref{sec:nudge}) redirects an arm extension already produced by the
generator, whereas RePointer's IK solver directly optimises forearm--target
alignment. Their different Score$_b$ values therefore reflect, in part, the
strength and constraints of these geometric operators rather than solely
learned scene-aware gesture generation. A harder future benchmark should
withhold the target coordinate and require systems to infer the referent from
the scene and utterance.

\paragraph*{Temporal evaluation requires an explicit tolerance.}
Unlike spatial grounding, gesture timing has no natural exact-frame target:
the reference provides one kinematically defined apex, while communicatively
acceptable timing may span an interval. We therefore use a fixed tolerance
for all clips and report raw timing errors alongside Score$_a$. The
RePointer--MM-Conv-Flow ordering is stable across a broad sensitivity sweep,
but absolute scores and the RePointer--GT STG comparison are not. The
tolerance should consequently be understood as part of the benchmark
definition rather than as an empirically established perceptual threshold.

\paragraph*{Geometric grounding and naturalness are complementary.}
RePointer substantially exceeds MM-Conv-Flow on spatial grounding,
but the naturalness study found no statistically significant
difference between the two generated systems.
Conversely, distributional motion
metrics also disagree with perception: RePointer obtains the lowest FGD and
highest velocity diversity and beat alignment without a corresponding MOS
advantage. No single objective score therefore captures referential gesture
quality. Future benchmarks should retain separate measures of geometric grounding, temporal alignment, and perceptual motion quality.

\section{Conclusion}
We introduced a benchmark for spatially grounded gesture generation: pointing-annotated clips from a naturalistic conversational dataset with 3D referents, a task in which systems must determine when and where to point within continuous co-speech motion, and a protocol that reports temporal alignment, spatial grounding, and perceived naturalness separately. Evaluating a flow-matching baseline, an independently developed retrieval-based system, and captured human motion shows why this separation matters: the retrieval-based system exceeds captured motion on spatial grounding and distributional motion metrics without being rated more natural. Geometric precision, timing, and perceived motion quality are therefore distinct task dimensions, and no single score captures referential gesture quality. The released data, baseline, and scorer provide a fixed reference point for future work on referential communication in embodied agents; a referent-identification study on a dedicated distractor-controlled target set is the next step toward measuring communicative success directly.

%% file: appendix-arxiv.inc
\appendix
\renewcommand{\thesection}{\Alph{section}}
\section{Temporal-Score Sensitivity}
\label{sec:supp-tau}

The temporal score defined in \S\ref{sec:metric} has two
design parameters: the tolerance $\tau$, within which apex-timing
errors receive full credit, and the Gaussian decay $\sigma_t$ outside
that tolerance. The primary evaluation uses $\tau=10$ and
$\sigma_t=15$ frames (0.33\,s and 0.5\,s). These values define the
benchmark score rather than estimating a human perceptual tolerance.
Here we test which conclusions depend on their choice.

\paragraph*{Setup.}
We evaluate a fine grid of $\tau\in\{0,\ldots,30\}$ and
$\sigma_t\in\{5,\ldots,30\}$ frames in one-frame increments, giving
806 parameter combinations. Detected and reference apexes and
Score$_b$ are fixed throughout; only the temporal-score
parameterisation changes. Confidence intervals use the same
recording-clustered bootstrap as in \S\ref{sec:objective-results} (59 source
recordings, 5,000 resamples).

\begin{figure}[t]
    \centering
    \includegraphics[width=\linewidth]{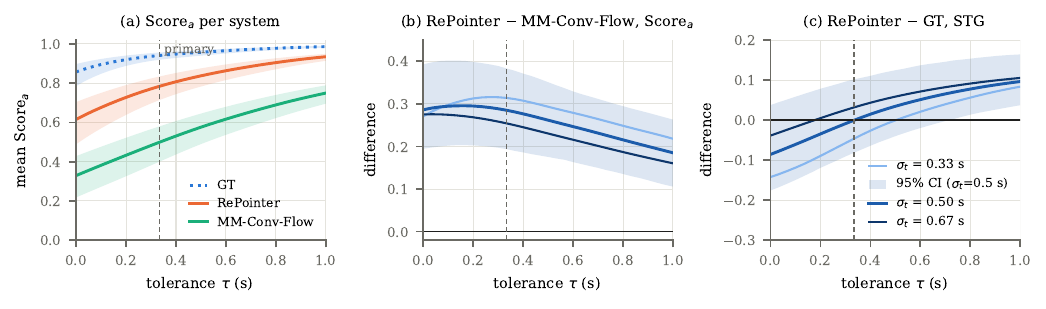}
    \caption{\textit{Sensitivity of the temporal evaluation to
    $\tau$ and $\sigma_t$.}
    (a) Mean Score$_a$ for GT, RePointer, and MM-Conv-Flow.
    Absolute score levels vary substantially with the temporal-score
    parameters.
    (b) Difference in Score$_a$ between RePointer and MM-Conv-Flow;
    RePointer remains higher throughout the tested parameter range.
    (c) Difference in STG between RePointer and GT; this ordering
    changes with the temporal-score parameters.
    The primary benchmark setting is $\tau=10$, $\sigma_t=15$ frames.}
    \label{fig:supp-tau-curves}
\end{figure}

\paragraph*{Absolute score levels are parameter-dependent.}
Score$_a$ increases monotonically with both $\tau$ and $\sigma_t$
for every system (Fig.~\ref{fig:supp-tau-curves}a). Across the grid,
mean Score$_a$ ranges from 0.30 to 0.97 for RePointer and from
0.11 to 0.84 for MM-Conv-Flow. Absolute Score$_a$ values therefore
depend strongly on the stated parameterisation. The raw median apex
errors reported in \S\ref{sec:objective-results}---0.40\,s for RePointer and
0.92\,s for MM-Conv-Flow---are independent of these parameters.

\paragraph*{RePointer versus MM-Conv-Flow is robust.}
RePointer obtains higher Score$_a$ than MM-Conv-Flow under all
806 tested parameter combinations
(Fig.~\ref{fig:supp-tau-curves}b). The mean difference ranges from
+0.13 to +0.34, and the smallest lower 95\% confidence bound across
the grid is +0.06. The combined STG comparison gives the same
ordering throughout the grid, with a smallest mean difference of
+0.21 and smallest lower 95\% bound of +0.14.

\paragraph*{GT provides a detector/reference diagnostic.}
When the same detector is applied to captured motion, GT obtains
higher Score$_a$ than RePointer throughout the tested grid. Because
the reference apex is itself kinematically defined, we treat this
comparison as a diagnostic of agreement between the detector and
reference definition rather than as an estimate of human timing
variability or a human-performance ceiling.

\begin{figure}[t]
    \centering
    \includegraphics[width=0.8\linewidth]{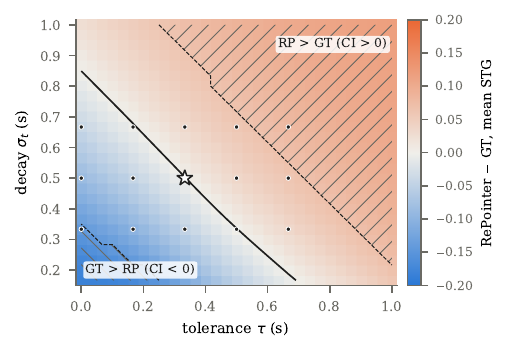}
    \caption{\textit{Sensitivity of the RePointer--GT
    combined-grounding comparison.}
    Each cell shows the mean STG difference (RePointer minus GT)
    for one $(\tau,\sigma_t)$ parameterisation. The black contour
    marks zero difference, hatching indicates regions where the
    95\% confidence interval excludes zero, and the star marks the
    primary setting $(\tau,\sigma_t)=(10,15)$. Its proximity to the
    zero contour illustrates why the RePointer--GT STG near-tie is
    parameter-sensitive.}
    \label{fig:supp-stg-signmap}
\end{figure}

\paragraph*{RePointer versus GT on combined grounding is
parameter-sensitive.}
The RePointer--GT comparison under STG behaves differently
(Figs.~\ref{fig:supp-tau-curves}c
and~\ref{fig:supp-stg-signmap}). RePointer has higher spatial
grounding than GT (0.905 versus 0.767) but larger apex-timing error.
Consequently, stricter temporal scoring gives greater weight to the
timing difference, whereas more lenient scoring increasingly exposes
RePointer's spatial advantage.

The sign of the mean RePointer--GT STG difference changes across the
grid. At $\sigma_t=15$ frames, it changes between $\tau=10$ and
$\tau=11$ frames, placing the primary setting close to the transition.
The 95\% confidence interval contains zero for 60\% of all 806
settings and for every setting in the neighbourhood
$\tau\in[5,15]$, $\sigma_t\in[10,20]$. Thus, the near-equality under
the primary setting (0.719 versus 0.720) is not a
parameter-independent comparison between RePointer and captured
motion.

\section{Apex Detector}
\label{sec:supp-apex}

The frozen apex detector is a logistic-regression classifier
($C=1$, balanced class weights), trained on recordings disjoint
from evaluation. Frames within $\pm3$ frames of the automatically
estimated reference apex are labelled positive; all other frames
are negative. Each frame is represented by 12 values obtained
through maximum and mean pooling of six kinematic features over
a $\pm10$-frame window. The $\pm3$-frame supervision window is
distinct from the scoring tolerance $\tau=10$ frames.
The sensitivity analysis above varies the scoring parameters
while keeping detector predictions fixed; it does not independently
validate apex detection.

\section{MM-Conv-Flow Reproducibility Details}
\label{sec:supp-mmconvflow}

This section provides implementation and training details for
MM-Conv-Flow beyond the architectural description in \S\ref{sec:baseline-flow}. Phase 1 (P1) trains the co-speech motion prior, while
Phase 2 (P2) adapts the model for spatially grounded pointing.

\subsection{Architecture Configuration}
\label{sec:supp-mmconvflow-architecture}

\begin{figure}[t]
    \centering
    \includegraphics[width=0.8\columnwidth]{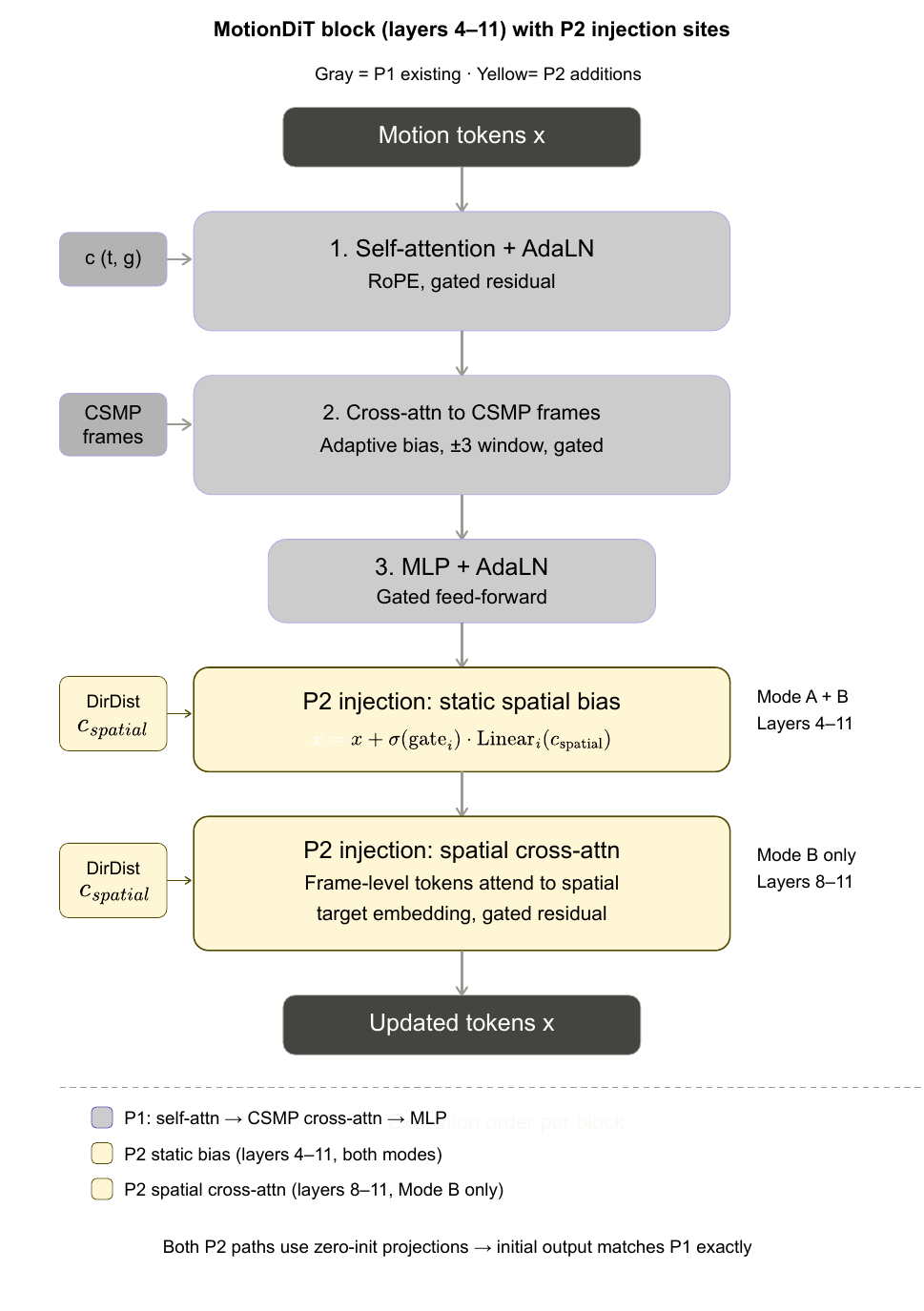}
    \caption{Single MotionDiT transformer block (layers 4--11) showing 
    P1 co-speech components (gray) and P2 spatial injection sites 
    (yellow). The spatial pathway adds a gated residual bias on layers 
    4--11 (Mode~A and B) and single-token cross-attention on layers 
    8--11 (Mode~B only). Both paths use zero-initialized projections, 
    so the initial P2 output matches P1 exactly.}
    \label{fig:supp-architecture}
\end{figure}

MM-Conv-Flow uses a 12-layer MotionDiT backbone
($d=768$, 12 heads) on 180-frame, 30-Hz SMPL-X sequences
(300-D/frame). Frozen 768-D data2vec audio/text representations
provide global and frame-level CSMP conditioning, with frame-level
attention restricted to $\pm3$ frames. P2 encodes target direction,
log-distance, and pointing hand into a 768-D representation. As
shown in Fig.~\ref{fig:supp-architecture}, spatial conditioning is
injected as a gated residual bias in layers 4--11 and, in Mode B,
through cross-attention in layers 8--11; the spatial projections are zero-initialized, the bias gates are initialized to $-2$, and the cross-attention gates to $-3$.

The P1 model contains 477.3M parameters, comprising the
170.5M-parameter MotionDiT backbone and the frozen
306.8M-parameter CSMP encoder. Mode B adds approximately 0.8M parameters
for spatial conditioning, yielding 478.1M parameters in total, of
which 171.3M are trainable.

\subsection{Training Configuration}
\label{sec:supp-mmconvflow-training}

P1 is trained on BEAT2 to learn the general co-speech motion prior.
P2 is initialized from the final P1 checkpoint and trained on a
70/30 mixture of pointing and BEAT2 samples. The BEAT2 component
retains general co-speech motion during spatial adaptation.
Table~\ref{tab:supp-training} gives the optimization settings.
\begin{table}[t]
\centering
\caption{Training configuration for the co-speech prior (P1) and
spatial adaptation (P2). A dash denotes a loss not used in that phase.}
\label{tab:supp-training}
\begin{tabular}{p{0.31\linewidth}p{0.27\linewidth}p{0.34\linewidth}}
\toprule
Setting & P1 & P2 \\
\midrule
Data & BEAT2 & 70\% pointing, 30\% BEAT2 \\
Optimizer & AdamW & AdamW \\
Learning rate
 & $2\times10^{-4}$
 & $3\times10^{-5}$ backbone;
   $5\times10^{-5}$ spatial \\
Weight decay & 0.05 & 0.01 \\
Effective batch size & 192 & 64 \\
Epochs & 1100 & 200 \\
Audio dropout & 0.15 & 0.30 \\
Text dropout & 0.15 & 0.10 \\
Spatial dropout & -- & 0.10 \\
EMA decay & 0.9999 & 0.9999 \\
\midrule
\multicolumn{3}{l}{\textit{Loss weights}} \\
Flow matching                & 1.0 & 1.0 \\
Total variation              & 0.5 & -- \\
Velocity                     & 1.0 & 0.01 \\
Jerk                         & 2.0 & 0.05 \\
Jerk-distribution matching   & 2.0 & -- \\
Bone length                  & 2.0 & -- \\
Relational                   & --  & 0.5 \\
\bottomrule
\end{tabular}
\end{table}

The total-variation, velocity, jerk, jerk-distribution, and
relational losses are weighted by $t^2$; bone length is not.
In P2, $\mathcal{L}_{\mathrm{rel}}
=1.5\mathcal{L}_{\mathrm{align}}
+0.5\mathcal{L}_{\mathrm{extend}}$ has overall weight 0.5, with
the component losses defined in \S\ref{sec:baseline-flow}.
P1 uses no relational or spatial loss; total variation,
jerk-distribution matching, and bone-length regularization are
disabled in P2.

\section{Auxiliary Motion Metrics}
\label{sec:supp-motion-metrics}

\paragraph*{Fr\'echet Gesture Distance (FGD).}
We use the EMAGE motion encoder to compare the feature
distribution of each evaluated system with the training-reference
distribution. Lower FGD indicates closer distributional agreement.

\paragraph*{Average jerk.}
Jerk measures the rate of change of acceleration. We report its
average magnitude to characterise motion smoothness. Lower values
indicate smoother motion, although smoothness alone does not imply
naturalness.

\paragraph*{L1 velocity diversity.}
Characterises variation in motion velocities using
L1 distances. Larger values indicate greater velocity variation,
which may reflect either richer movement or excessive motion.
We therefore assign no preferred direction to this metric.

\paragraph*{Beat alignment.}
Beat alignment measures temporal correspondence between motion
beats and audio beats. Higher values indicate stronger alignment.
This captures rhythmic audio--motion correspondence, whereas
Score$_a$ evaluates the timing of the referential apex.

\section{Supplementary Video}
\label{sec:supp-video}
A video showing GT MoCap, MM-Conv-Flow and RePointer side by side
(left to right) on five held-out test clips (IDs as in Fig.~\ref{fig:mos}):
181, 88, 115, 298 and 135, is available at
\url{https://huggingface.co/datasets/hsi-workshop/referential-gesture-challenge}.
All conditions share the same speech audio, avatar, room and camera;
only body motion differs. Clip 298 illustrates the retrieval failure
mode discussed in \S\ref{sec:validity}.